# A Hybrid Model for Stock Market Forecasting: Integrating News Sentiment and Time Series Data with Graph Neural Networks

**Nader Sadek, Mirette Moawad, Christina Naguib and Mariam Elzahaby**

Faculty of Computers and Artificial Intelligence Cairo University, Egypt

nader.zaki08@gmail.com
miretteamin@gmail.com
ch.saad825@gmail.com
elzahabymariam@gmail.com

**Abstract:** Stock market prediction has long been a challenging problem in the field of finance and investment. Accurately predicting the movements of stock prices is crucial for making informed decisions and maximizing investment returns. Traditional models mainly use historical prices. We found that there is a gap in research in integrating financial news into the model, which has emerged as a promising direction in enhancing predictive accuracy. This research aims to address this problem by exploring a multimodal approach by combining companies' news articles and their historical stock data to predict future stock movements. The objective was to compare the performance of a Graph Neural Network (GNN) model with an LSTM model. The methodology employed in this research involves an LSTM model that embeds the historical data for each company and a language model to embed news articles. These embeddings will represent nodes that have relationships presented by edges within a graph. Using a GNN message aggregation technique known as GraphSAGE, the model should be able to capture interactions and dependencies between news articles, companies, and industries and use this information to predict future stock movements. Two target variable approaches are explored: one focusing on the binary classification of whether the stock price will increase or decrease, and the other considering the significance of the increase. This methodology was evaluated on two datasets, the US equities dataset and the Bloomberg dataset. The results showed that the GNN model was able to achieve better performance than the baseline LSTM model on both datasets. The GNN model achieved an accuracy of 53% on the first target, a statistically significant 1% improvement over the baseline, and a 4% precision gain on the second target, which confirms the effectiveness of exploiting financial news using graph-based models. Furthermore, we observed that increasing the number of news samples led to improved accuracy. We also find that headlines contain stronger predictive signal than full articles which is consistent with evidence that headlines disproportionately shape readers' judgments and market reactions.

**Keywords**: Deep learning, Graph neural networks, Finance, Knowledge graph

## 1. Introduction

The stock market continues to play a crucial role in the global economy, providing a platform for individuals to invest their savings and potentially earn returns. Despite recent fluctuations, the stock market remains a key indicator of the overall health and stability of a nation's economy, and it continues to shape the economic landscape and investment decisions of both individuals and organizations. Therefore, financial analysts use various types of analysis for stock evaluation and decision-making; however, our focus is on both "Technical Analysis" (Fama & French, 1992) and "News-based Analysis". By definition, technical analysis uses charts and other statistical analysis to identify patterns in a stock's price and volume data to predict future price movements while news-based analysis involves evaluating the potential impact of news events, such as earnings reports, product launches, regulatory changes, and other relevant events, on a company's stock price and overall financial performance. AI has the potential to revolutionize the stock market prediction process, in which a lot of methods that use technical analysis or news-based analysis have been evaluated, whereas combining both approaches and representing them in a knowledge graph has not yet been extensively explored. Consequently, we have chosen GNNs that can mimic human behaviour and are particularly well-suited to tasks that involve modeling relationships and interactions between entities in a graph, such as those found in financial markets. But the stock market is a complex system, and making predictions about it is a challenging task. Hence, we will investigate whether using the graph representation helps the model learn how external information influences individual companies by considering their industry context and news impact.

## 2. Background and Related Work

In this section, we will provide an overview of the key components and techniques utilized in our model. And would also provide an overview of previous studies or works that have used similar techniques to predict the stock market. Our objective is to contextualize the research by highlighting the gap in the literature that the current study aims to fill and to provide evidence of the novelty of the current research. The following papers will review studies that have used GNNs for stock market prediction, critically evaluate their methodologies and results, and explain how they differ from the current study.

## 2.1 Background

### 2.1.1 Graph neural networks

Let a graph be defined as $G = (V, E)$ where:

- $V$ is the set of nodes, with $|V| = n$.
- $E \subseteq V \times V$ is the set of edges, with $|E| = m$.

The graph can be represented by an adjacency matrix $A \in \{0, 1\}^{n \times n}$, where:

$$A_{ij} = \begin{cases} 1 \; if \; (i,j) \in E \\ 0 \; otherwise \end{cases}$$

Each node $i \in V$ has a feature vector $x_i \in R^d$, and all node features are collected in a matrix $X \in R^{n \times d}$

The task is to predict a label $y_i \in \{0, 1\}$ for each node $i$.

The GNN follows a message-passing framework where, at each layer $k$, messages from neighboring nodes are aggregated and then combined with the current node's embedding. The message passing update rules at layer $k$ are given by:

$$m_i^{(k)} = AGG^{(k)}\left(\left\{\left(x_j^{(k-1)}, x_i^{(k-1)}\right) : (i,j) \in E\right\}\right) \quad (1)$$

Next, the embedding of node $i$ is updated as follows:

$$x_i^{(k)} = COM^{(k)}\left(x_i^{(k-1)}, m_i^{(k)}\right) \quad (2)$$

- $m_i^{(k)}$: The aggregated message for node $i$ at layer $k$, which is a function of the neighboring nodes $j$ and their respective embeddings $x_j^{(k-1)}$ from the previous layer.
- $AGG^{(k)}(\cdot)$: The aggregation function at layer $k$, which aggregates the messages from the neighbors of node $i$.
- $x_i^{(k)}$: The new embedding for node $i$ at layer $k$, obtained by combining the current embedding
- $x_i^{(k-1)}$ with the aggregated message $m_i^{(k)}$.
- $COM^{(k)}(\cdot)$: The combination function at layer $k$, which combines the current node embedding $x_i^{(k-1)}$ with the aggregated message $m_i^{(k)}$.

### 2.1.2 Financial text analysis

FinBERT (Yang, et al., 2020) is a pre-trained language model that is specifically designed for financial text analysis. It is built by further training the popular Bidirectional Encoder Representations from Transformers (BERT) (Devlin, et al., 2018) architecture on a large corpus of financial documents, including SEC filings, earnings call transcripts, and news articles. Then it was fine-tuned for financial sentiment classification. It prepends a special token to the sentence [CLS]. In our context, the mentioned model was used to extract the news embedding.

### 2.1.3 Time series analysis

Time series analysis is a statistical method used to analyze data collected over time. In our context, time series analysis is used to analyze historical stock price data to identify patterns and trends, to make predictions about future stock prices, and to extract company embeddings using LSTM and Transformer models (Vaswani, et al., 2017).

## 2.2 Literature Review

In this section, we provide an overview of previous studies that have employed similar techniques for stock market prediction to contextualize our research and demonstrate its novelty. The existing body of work can be grouped into several key approaches based on the data used to construct the graph relationships.

One cluster of research focuses primarily on graphs derived from textual news data. For example, the *Heterogeneous graph knowledge enhanced stock market prediction* paper (Xiong, et al., 2021) which focused on predicting stock price fluctuations based on financial text using a Multi-grained Heterogeneous Graph (HGM-GIF) framework. A key limitation, unlike our approach, is that they solely relied on news data and did not incorporate time series data. Similarly, the *Modeling the Stock Relation with Graph Network for Overnight Stock Movement Prediction* paper (Li, et al., 2020) introduced an LSTM Relational Graph Convolutional Network (LSTM-RGCN) model to predict overnight stock movements based on news titles. While these methods are adept at capturing event-driven relationships, they overlook the quantitative patterns found in historical stock prices.

A second approach involves constructing graphs from market data or pre-defined external knowledge, without incorporating news. The paper *Exploring Graph Neural Networks for Stock Market Predictions with Rolling Window Analysis* (Matsunaga, et al., 2019) employed Temporal Graph Convolution (TGC) to predict long-term stock values using supplier interaction data but did not utilize news information. Additionally, the *Using External Knowledge for Financial Event Prediction Based on Graph Neural Networks* paper (Yang, et al., 2019) focused on event prediction rather than stock movements, using graph neural networks to forecast future corporate events based on structured knowledge graphs. Along these lines, a recent paper builds a dynamic graph based on the similarity of stock features every day; however, it also doesn't consider information outside stock parameters and makes connections between companies based only on stock prices, which means it doesn't use news. These models effectively capture market structure but miss the crucial context provided by external news events. Taking this concept further, the paper Stock market price forecasting using evolving graph neural network (Korablyov, et al., 2025) propose an "evolving graph neural network" that uses a dendritic artificial immune network (DaiNet) to dynamically cluster stocks and update the graph structure daily based on price movements. While these dynamic methods are sophisticated, their relational graphs are still derived internally from price correlations, and they miss the crucial external context provided by financial news events.

Finally, some studies attempt to fuse multiple data sources. The paper A graph neural network-based stock forecasting method utilizing multi-source heterogeneous data fusion (Li, et al., 2022) developed a comprehensive model that fused transaction data, news text, and graphical indicators. In *Combining Financial Data and News Articles for Stock Price Movement Prediction Using Large Language Model* (Elahi & Taghvaei, 2024) used news articles with few-shot learning on large language models to predict stock movement, but they did not use graph structures and evaluated only 20 companies. While these approaches show promise, the integration of both news-based relationships and historical time-series data within a single GNN framework remains underexplored, which is the primary gap our research addresses. More recently, several survey and review papers have examined the broader use of Graph Neural Networks in financial prediction tasks. For instance, *Graph Neural Networks for Financial Market Prediction: Challenges, Opportunities, and Future Directions* (Zhou, et al., 2024) provides a comprehensive overview of existing architectures, datasets, and methodological challenges. This work further supports the relevance of integrating multimodal signals, such as market structure and textual sentiment, which aligns closely with our hybrid modeling approach. Our approach is distinct in its focus on using news to define the relationships between companies while using their historical stock data as the core features for prediction. This review highlights the unique contributions of our research in integrating news sentiment analysis and time series data for stock market prediction.

## 3. Task Formulation and Dataset

### 3.1 Task

Our goal is to predict the future stock price reaction based on the given news articles and stock history related to each company. We formulate the prediction as a classification problem in which the target class will be "1" for the stock price increase and "0" for the stock price decrease.

In our model, we explored two different approaches during the training process. The first approach involved extracting a single target variable, which focused on whether the stock increased. This binary classification task aimed to predict whether the stock price would go up or down without considering the magnitude of the change.

The target variable $y_t$ is defined as follows:

$$y_t = \begin{cases} 1, & \text{if}(\text{close}_t - \text{close}_{t-1}) > 0) \\ 0, & \text{otherwise} \end{cases}$$

Where:

- $close_t$ represents the stock's closing price at time $t$.
- $close_{t-1}$ represents the previous closing price.
- $y_t = 1$ if the stock price has increased, and $y_t = 0$ if it has decreased or remained the same.

This approach focuses on predicting the direction of stock price movement, without considering the magnitude of the change.

However, inspired by the relevant research paper Modeling the Stock Relation with Graph Network for Overnight Stock Movement Prediction [Li et al., 2020], and considering real-world investment scenarios in which investors prefer significant returns rather than marginal ones. Therefore, we decided to adopt a second approach. In this approach, we introduced an additional target variable that accounted for the significance of the stock increase. By incorporating this variable, our model aimed to predict not only if the stock would increase but also whether the increase would be significant or not. Yet, we observed that [Li et al., 2020] used this significance-based target and only used accuracy as its evaluation metric, which seems to ignore the fact that using a shifted threshold with such data can alter the reported accuracy and potentially obscure the model's performance. as it can constantly predict one class and result in a deceptively higher accuracy. We avoided such problem by incorporating other metrics like F1, Precision, and Recall alongside accuracy.

$$y_t = \begin{cases} 1, & if(\text{close}_t - \text{close}_{t-1}) > 0.04 \times \text{std}(\text{close}_{t-N:t}) \\ 0, & \text{otherwise} \end{cases}$$

Where:

- $\text{std}(\text{close}_{t-N:t})$ is the standard deviation of the previous $N$ closing prices *(the look back window)*.
- The factor 0.04 scales the threshold to represent a 4 % volatility-adjusted change.
- In our experiments, $N = 100$, i.e., we used the last 100 closing prices to compute this rolling standard deviation.

### 3.2 Dataset

We conducted experiments on two news datasets covering different time intervals and different sets of stocks to provide stronger evidence in favor of GNNs and LSTM models and compare their efficacy.

The US equities dataset includes 82k articles written between 2015 and 2020 related to 325 companies that trade on NASDAQ (Nasdaq, n.d.). We split it into 1100 training days and 175 testing days.

The Bloomberg dataset includes 450k articles from 2006 to 2013 on 625 companies that trade on NASDAQ and NYSE. Here, we drop the first 2 years as they include very few articles. And we split it into 748 training days and 117 testing days.

## 4. Methodology

### 4.1 Data Preprocessing

In this section, we outline the steps taken to preprocess our default dataset "US equities" before training the predictive model.

#### 4.1.1 Companies extraction from news

To ensure that the news articles were relevant to the Nasdaq market, we extracted company names from the Nasdaq screener dataset using a customized named entity recognition (NER) approach. We then filtered both the news and historical data to exclude any data points that were not associated with the Nasdaq market.

#### 4.1.2 Datasets alignment

The extracted company names were used to create a generic version of companies' names by mapping them to their corresponding symbols. This step was taken to ensure consistency in the representation of company names across different data sources and to facilitate the integration of different types of financial data. We also aligned the dates presented in the news dataset and historical dataset.

#### 4.1.3 News encodings

To obtain news embeddings for the news article titles, we utilized a financial embedding model known as Sigma (Sigma AI, 2022), then we saved the news vector extracted to be used in graph creation later.

The Sigma model is a financial sentiment analysis tool built on a transformer architecture like BERT, tailored for finance-specific data. It was fine-tuned using financial documents like SEC filings, earnings call transcripts, and news articles. In this study, Sigma extracts embeddings from financial text, encoding sentiment and context into vector representations. These embeddings help in understanding the influence of news on stock price predictions, as the model learns from the financial relevance captured by Sigma.

### 4.2 Graph Construction

The graph used in this study consists of two primary components: nodes and edges. Nodes can represent a company, an industry, or an article, while edges denote the relationships between nodes. Specifically, each article node connects to the main company and other mentioned companies that may impact the main company in some way. Additionally, each company node is associated with a particular industry. Each type of node in the graph has its own set of features. The article nodes are represented by news node embeddings, while the company nodes are characterized by concatenated features of historical stock embeddings and learnable embeddings. The industry nodes are represented solely by learnable embeddings. These embeddings are generated by randomly initialized weights, which are then updated while training the GNN. We built the training graph dataset for 15 days before the target stock day. To further ensure that GNN is better than LSTM only, we used the Bloomberg dataset. As this dataset is different from the US equities that we showed its graph creation for above, the graph changed as we eliminated the industry node and extracted the news embedding differently by splitting the article into sentences and then passing them to the Sigma model to get the news node embedding.

### 4.3 Proposed Model Architecture

The historical stock data used in this study consisted of five main features: open, high, low, close, and volume. We used the close price feature of companies' time series to obtain company embeddings encoded by an LSTM model. This LSTM model takes a sequence of 15-time steps (days). These 15 days window was chosen based on recommendations from financial analysts, as it is considered an optimal period to capture meaningful short-term market trends and momentum while filtering out the noise of daily price volatility. Shorter periods often contain too much random fluctuation, whereas longer periods might smooth out more immediate, actionable signals, making 15 days a balanced choice for modeling significant price movements.

The model processes these sequences through a batch normalization layer first; we didn't normalize the historical data beforehand because we were experimenting with different data-splitting strategies to simplify data handling without compromising model evaluation by causing information leaks. The LSTM consists of 2 layers, and its activations are then passed as company node embeddings for the GNN model. Therefore, the LSTM layers are trained jointly with the GNN, for which we used GraphSAGE as the message-passing layer.

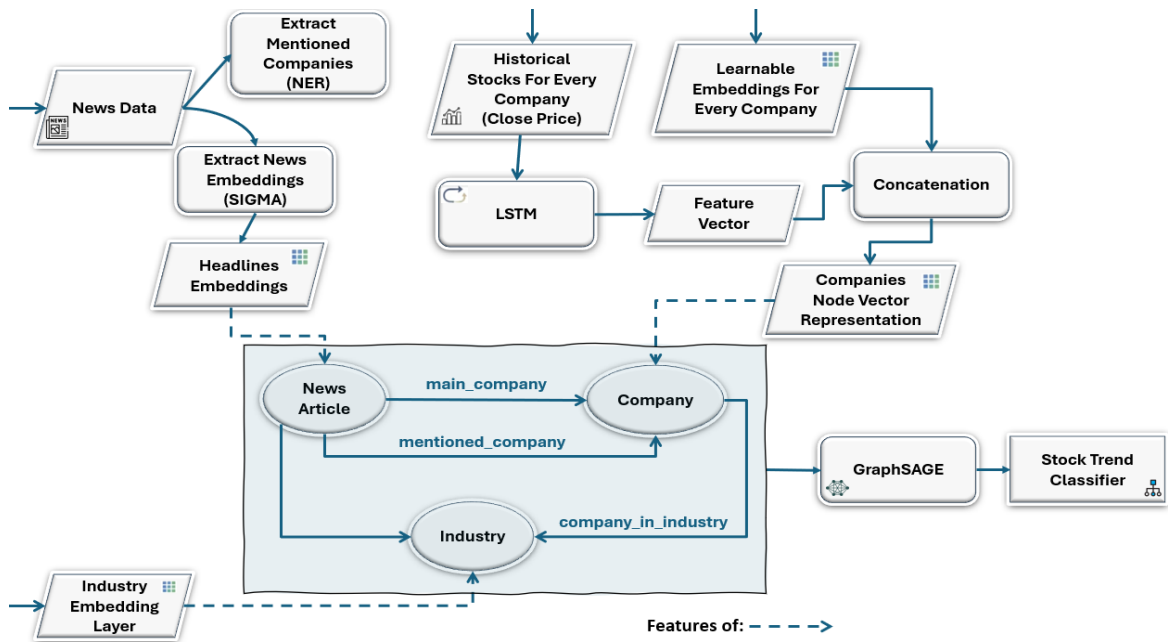

**Figure 1: Model Architecture**

## 5. Results and Analysis

This section presents the experiments we went through and our final model results analysis, as our objective was to prove that the LSTM model will not be as effective as the GNN model. Therefore, our baseline model is the LSTM model only trained on historical data, specifically the close price for each company. Then, we introduced the GNN model for the classification problem with the 2 approaches for the target labels that we mentioned before. First, we conducted some experiments on GraphSAGE conv and chose the best out of them to explore the performance of other GNN techniques and exclude the weak ones. After each experiment, we are going to present every experiment's evaluation using 2 target definitions' results and show the different metrics that are explained in the methodology chapter. The metric for the first target definition is accuracy, as the data is balanced, while the metric for the other target definition is precision, as the data became unbalanced by extracting only a significant increase.

### 5.1 Experiments

In this section, we will present the experimental models we built in detail, along with their results. For each experiment, we calculate the following metrics:

- Accuracy calculated for the model trained with the first target definition.
- Precision calculated for the model trained with the second target definition.
- Both accuracy and precision on 100 companies for which the model gives predictions with the highest confidence (highest predicted probability).

Any model used in the experiments is explained in the background section 2.

All experiments were conducted on an NVIDIA RTX 2070 GPU, and all the experiments' configurations are as follows:

**Table 1: Configuration Table for Both Targets Across Experiments**

| *Hyperparameters* | *US Equities* | *Bloomberg* |
|---|---|---|
| **Number of Epochs** | 55 | 55 |
| **Learning Rate** | 0.0001 | 0.0001 |
| **Node Embedding Dimension** | 128 | 128 |
| **Optimizer** | AdamW | AdamW |
| **GNN Layers' Number** | 3 | 3 |
| **Batch Size** | 4 | 4 |
| **Highest Confidence Companies** | 100 | 200 |

*5.1.1  Baseline model*

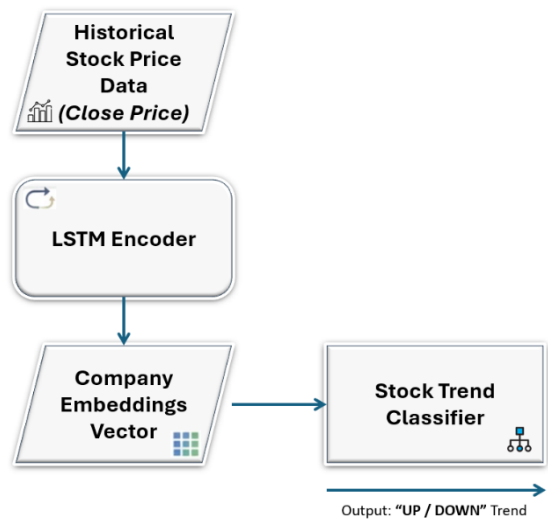

**Figure 2: Baseline Model Architecture**

This experiment represents the baseline model, in which we used the close price feature of companies' time series as an input for the LSTM model to predict the stock movement.

*5.1.2 Experiments' results table*

We changed a component with each experiment within the 4 components Node Aggregation Method, News Encoding Model, Timeseries Encoding and which news section to encode. The table below presents all experiments results knowing that best results are shown in **bold**, and the weakest results are in *italics*.

**Table 2: Results of each experiment**

| *Experiment* | *Aggregation* | *News Model* | *Timeseries Encoding* | *News Section* | *Acc 1st Target ↑* | *Precision 2nd Target ↑* | *Acc@High Certainty ↑* | *Precision@High Certainty ↑* |
|---|---|---|---|---|---|---|---|---|
| *1* | GraphSage | Sigma | LSTM Close Price | Article Content | 53% | 53.4% | 54.2% | 53% |
| *2* | GraphSage | Sigma | LSTM Close Price | Article Title | 53% | 55% | 54.2% | 53% |
| *3* | GAT | Sigma | LSTM Close Price | Article Title | 52% | 46% | 52.63% | 25.14% |
| *4* | GraphSage | FinBERT | LSTM Close Price | Article Title | 52.9% | 54% | 54.14% | 53.2% |
| *5* | GraphSage | Sigma | LSTM RSI | Article Title | 53% | 54.14% | 54.5% | 57% |
| *6* | GraphSage | Sigma | Transformer SMA | Article Title | 52.94% | 53.5% | 54.3% | 54.9% |

*5.1.3 Analysis of experimental configurations*

As observed in **Table 2: Results of each experiment**, different model configurations yielded varying results. Our analysis provides several insights into these performance differences.

*Note: Experiments 1 vs. 2 isolate headlines vs. content effects; titles yield higher precision.*

**GraphSAGE vs. GAT: Experiment 3**, which used a Graph Attention Network (GAT), showed a significant drop in performance, especially in precision for the second target. This can be attributed to the nature of our graph construction. GATs typically excel on homophilic graphs, where connected nodes are likely to share the same class (e.g., friends in a social network). We constructed our graph by connecting companies mentioned together in the news, which does not guarantee they will have similar stock movements. GraphSAGE, with its neighborhood sampling and diverse aggregation functions, is less reliant on the homophily assumption and thus performed more robustly on our heterogeneous graph.

This observation aligns with findings from recent studies showing that attention-based GNNs such as GATs tend to perform best on homophilic graphs, while heterophilic settings, where connected nodes often have dissimilar labels, favor architectures like GraphSAGE that aggregate more diverse neighbourhood information (Zhu, et al., 2020) (Pei, et al., 2020).

**Sigma vs. FinBERT:** We also compared news encoding models. **Experiment 2 (Sigma)** slightly outperformed **Experiment 4 (FinBERT)**. This is likely because Sigma is a version of FinancialBERT that has been fine-tuned on a larger and more domain-specific corpus of financial documents, allowing it to capture the nuances of financial news more effectively.

**Headlines vs. full text. Experiments 1 and 2** differ only in the news section encoded. Titles consistently matched or outperformed full-content variants (e.g., Precision 55 % vs. 53.4 %), suggesting that headlines capture the most behaviourally relevant information for short-term market reactions. This agrees with prior behavioural and neurocognitive evidence that headlines alone can bias readers' interpretations and investor sentiment (Baum, et al., 2021) (Hirshleifer, et al., 2023).

*5.1.4 Statistical significance (one-sample proportion Z-test)*

We test whether the observed accuracy exceeds random chance (balanced binary task).
**H₀:** $p = 0.50$ (random)   **H₁:** $p > 0.50$.
**Observed:** $\hat{p} = 0.53$, $N = 56{,}875$ (325 companies × 175 days).

**Standard error:** $\text{SE} = \sqrt{\frac{p_0(1-p_0)}{N}} = \sqrt{\frac{0.5 \cdot 0.5}{56{,}875}} \approx 0.00210$.

**Z-score:** $Z = \frac{\hat{p}-p_0}{\text{SE}} = \frac{0.53-0.50}{0.00210} \approx 14.31$.

With $\alpha = 0.05$, the one-sided critical value is $Z_{0.95} = 1.645$ (two-sided 1.96). Our result is highly significant (p $\ll 0.001$), so we reject $H_0$ and conclude the model performs better than chance. (Methodology per standard one-sample proportion z-test).

### 5.2 Final Model Results Analysis

In this section, we will present some analysis conducted on the best and final model we trained in terms of accuracy and precision which you can find in section **Table 2: Results of each experiment**, **Experiment 2** and its results are shown in the following Table 5-3 GraphSAGE Experiment 2 Results. Our goal is to investigate model behavior, its utility, and how it can be improved. The following chart Figure 5-10 reflects the mean number of articles mentioned in the top 50 accuracy companies and in the lowest 50 accuracy companies. Overall, the pattern between accuracy and number of articles is highlighted; when the number of edges between articles and companies increases, the accuracy is positively affected. Hence, it implies the effectiveness of including news in our study; in addition to, representing data in a graphical format which keeps a meaningful representation between articles and companies but also differs between the "mentioned companies" and "main company" edges.

**Table 3: Model results on the test set for each target compared to baseline**

|  | *Model* | *Accuracy ↑* | *Precision ↑* | *Recall ↑* | *F1 Score ↑* |
|---|---|---|---|---|---|
| *Default Target* | LSTM Baseline | 0.52 | 0.53 | 0.63 | 0.57 |
|  | Proposed GNN Model | **0.53** | 0.527 | **0.72** | **0.61** |
| *Target with Significance* | LSTM Baseline | 0.54 | 0.51 | 0.003 | 0.006 |
|  | Proposed GNN Model | **0.546** | **0.555** | **0.02** | **0.04** |

As shown in **Table 3: Model results on the test set for each target compared to baseline**, our proposed GNN model delivered an improvement over the LSTM baseline, achieving a 53% accuracy and a significantly higher recall. While the 1% absolute gain in accuracy might seem modest, this figure is a direct result of highly variable news data density across the dataset.

GNN's primary advantage is its ability to interpret the relational information embedded in financial news. However, many companies in our dataset had sparse news coverage on any given day. For these "news-poor" companies, the GNN could not leverage its unique graph-based capabilities, causing its performance to default to a level similar to the time series only LSTM.

Consequently, the overall 1% improvement masks a more significant effect: it represents a blended result where the strong predictive accuracy for news-rich companies is diluted by the baseline performance on companies lacking sufficient news data. This crucial relationship is confirmed by the analysis in **Figure 3: The Mean and Median of High and Low accuracy companies on the test set of US equities.**, which demonstrates a clear positive correlation between the number of news mentions and predictive success.

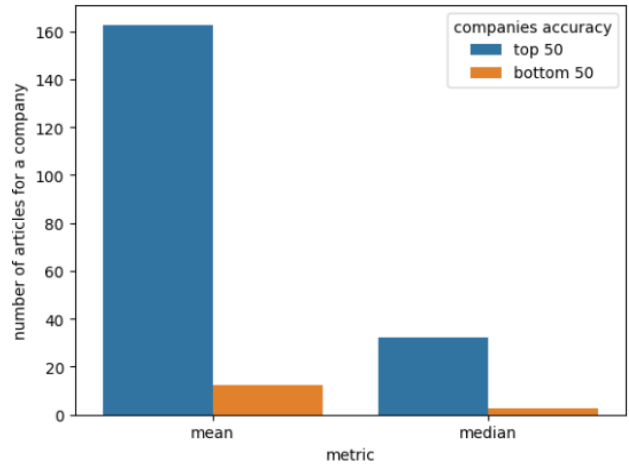

**Figure 2: The Mean and Median of High and Low accuracy companies on the test set of US equities.**

This bar plot shows that high-accuracy companies are mentioned much more frequently in articles than lower-accuracy companies, but then we checked the histograms of both sets of companies.

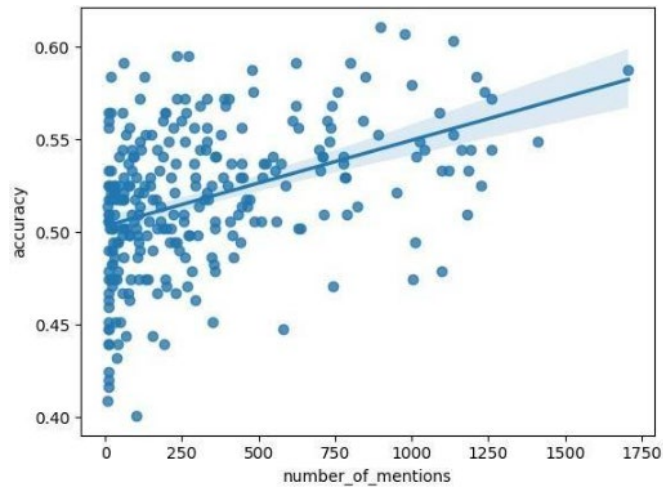

**Figure 3: The relationship between the number of mentions and model accuracy on the US equities dataset.**

Based on the previous plots, to improve the model, we can collect more news about other companies.

Our dataset time period was between 2015 and 2020, which induced some doubts about the behavior of the trained model, since the world witnessed significant fluctuations since 2019 due to COVID-19.

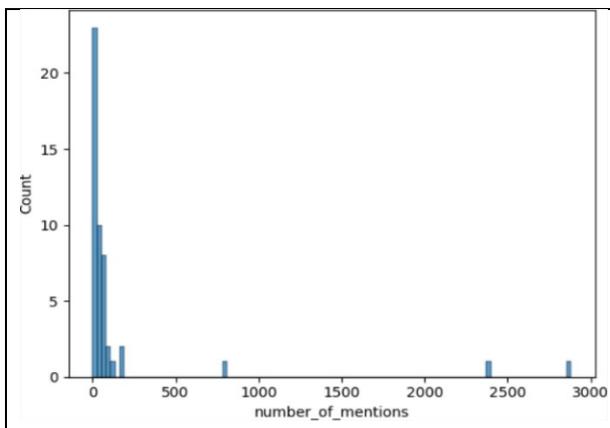

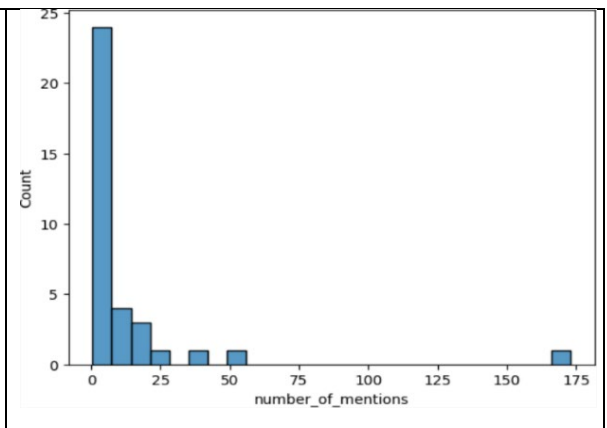

| **Figure 4: High Accuracy Companies Number of mentions Histogram** | **Figure 5: Low Accuracy Companies Number of mentions** |
|---|---|

## 6. Conclusion and Recommendation

In conclusion, our research has demonstrated that a Graph Neural Network, which integrates financial news relationships with historical time-series data, consistently outperforms a baseline LSTM model for stock market prediction. The GNN's ability to model the complex interplay between entities in the financial ecosystem confirms the value of this approach.

However, our analysis also highlighted several limitations of the current model, which in turn suggest clear directions for future work to enhance its performance and practical utility.

For practical implications the user of the model can focus on companies that are mentioned more in the news or identify the current set of high accuracy companies.

### 6.1   Limitation: Sensitivity to Textual Noise and the Need for Event Extraction

A key limitation observed during our experiments was the model's sensitivity to the signal-to-noise ratio in the input text. This was evident when using concise article titles proved to be more effective than using the full article content, which often contains significant irrelevant information. To address this, we recommend focusing on Event Extraction. By using advanced NLP techniques to isolate core financial events (e.g., mergers, earnings reports, product launches) from the article body, we could create a more potent and distilled textual representation. This would make the content more concise and could yield performance comparable to, or even better than, that obtained using only titles.

### 6.2   Limitation: Simplistic Graph Schema and the Opportunity for Competitor Data

While our graphical representation successfully captured inter-company relationships, its schema is inherently simplistic. The current graph connects companies based on co-mention, treating all relationships as equal without differentiating their nature (e.g., partnership vs. rivalry). To enhance performance, the graph schema could be modified to include richer semantic edges. For instance, if data on companies' strategic relationships, such as competitors and cooperators, were available, a distinct edge type could be incorporated between them. This would allow the model to learn more nuanced patterns of market influence.

### 6.3   Limitation: Lack of Robust Uncertainty for Practical Decision-Making

From a practical standpoint, the most significant limitation is that for making important investment decisions, a prediction alone is insufficient without a reliable measure of its confidence. While the predicted probability from a neural network offers a proxy for confidence, it is not a robust measure of model uncertainty. To bridge this gap for real-world application, we recommend Measuring Uncertainty in Model Predictions. A powerful approach would be to use the GNN for representation learning and feed these learned representations into a model explicitly designed to handle uncertainty, such as a Gaussian Process (GP) (Williams & Rasmussen, 2006). This would provide not just a prediction but also a confidence interval, creating a far more valuable tool for risk-managed investing.

## Acknowledgments and Disclosure of Funding

We would like to express our sincere gratitude to Dr. Cherry Ahmed (Faculty of Computers and Artificial Intelligence, Cairo University) for supervising this project and for her insightful guidance during its initial phases. We also thank the Faculty of Computers and Artificial Intelligence, Cairo University, for providing a supportive academic environment that nurtured our research. We are grateful to the anonymous ICAIR reviewers for their thoughtful feedback and constructive suggestions, which substantially strengthened this manuscript. We acknowledge the use of publicly available resources and tools, including the Bloomberg financial news dataset (used with permission from its author via the project repository), as well as Sigma and FinBERT for financial text embeddings. This research received no external funding; all experiments were conducted on personal computing resources.

**Ethics Declaration**: The bloomberg data we used required permission from premy.enseirb@gmail.com as mentioned on his GitHub repo: https://github.com/philipperemy/financial-news-dataset and we obtained his permission on Fri, Apr 28, 2023.

**AI Declaration**: We used AI tools to rephrase the abstract and ensure correct grammar.